\documentclass[conference,letterpaper]{IEEEtran}
\usepackage{graphicx}
\ifCLASSINFOpdf
\else
\fi
\usepackage{tabularx}
\usepackage{url}
\usepackage{subfig}
\usepackage{caption}
\usepackage{amsmath}
\usepackage{caption}

\begin{document}
\IEEEoverridecommandlockouts
\title{Scene-Specific Pedestrian Detection Based on Parallel Vision}
\author{
	\IEEEauthorblockN{Wenwen Zhang,~Kunfeng Wang,~\IEEEmembership{Member,~IEEE},~Hua Qu,~Jihong Zhao, and~Fei-Yue Wang,~\IEEEmembership{Fellow,~IEEE}}
\thanks{This work was partly supported by National Natural Science Foundation of China under Grant 61533019, Grant 71232006, and Grant 91520301.}
\thanks{Wenwen Zhang is with School of Software Engineering, Xi'an Jiaotong University, Xi'an 710049, China, and also with The State Key Laboratory for Management and Control of Complex Systems, Institute of Automation, Chinese Academy of Sciences, Beijing 100190, China (e-mail: zhangwenwen1000@gmail.com).}
\thanks{Kunfeng Wang (\emph{Corresponding author}) is with The State Key Laboratory for Management and Control of Complex Systems, Institute of Automation, Chinese Academy of Sciences, Beijing 100190, China, and also with Qingdao Academy of Intelligent Industries, Qingdao 266000, China (e-mail: kunfeng.wang@ia.ac.cn).}%
\thanks{Hua Qu is with School of Software Engineering, Xi'an Jiaotong University, Xi'an 710049, China.}
\thanks{Jihong Zhao is with Department of Communication and Information Engineering, Xi'an Jiaotong University, Xi'an 710049, China.}
\thanks{Fei-Yue Wang is with The State Key Laboratory for Management and Control of Complex Systems, Institute of Automation, Chinese Academy of Sciences, Beijing 100190, China, and also with the Research Center for Computational Experiments and Parallel Systems Technology, National University of Defense Technology, Changsha 410073, China (e-mail: feiyue@gmail.com).}
}

\maketitle

\begin{abstract}
 As a special type of object detection, pedestrian detection in generic scenes has made a significant progress trained with large   amounts of  labeled training data manually.
While   the models trained with generic dataset work bad when they are directly used in specific scenes.
 With special  viewpoints, flow light and backgrounds,   datasets from   specific scenes are much different from  the   datasets from generic scenes.
 In order to make the generic scene pedestrian detectors work well in specific scenes,
the  labeled  data   from  specific scenes are needed to adapt the models to the specific scenes.
 While   labeling the data manually spends much time and money, especially for specific scenes, each time with a new specific scene,   large amounts of images   must be labeled. What's more,  the labeling information is not so accurate in the pixels manually and different people make different labeling information.
In this paper, we propose an ACP-based  method, with augmented reality's help, we build the virtual world of  specific scenes, and make people walking in the virtual  scenes where it is possible for them to appear  to solve this problem of lacking labeled data and the results show that   data from   virtual world   is helpful to adapt generic pedestrian detectors to specific scenes. 

\end{abstract}

\IEEEpeerreviewmaketitle
\section{Introduction}
\begin{figure}[t]
	\centering
	\includegraphics[width=0.95\linewidth]{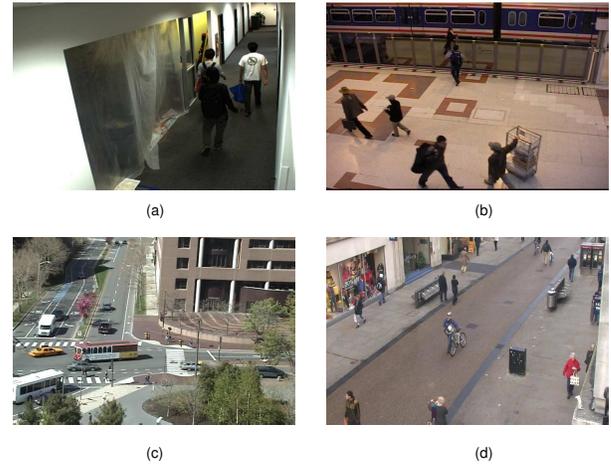}
	\captionsetup{font={footnotesize}}
    \caption{Examples of  surveillance equipments in specific scenes: \textbf{(a)} CMU indoor vision research \textbf{(b)} sub-open scene-subway station \textbf{(c)} MIT surveillance dataset from traffic scene \textbf{(d)} Town center scene.}
	\label{fig:figure1}
\end{figure}

Detecting pedestrians from    video sequences  in specific scenes  is very important to surveillance of traffic scenes and other public scenes such as  squares or super  markets \cite{wang2016multi}. It can be used to  analyze abnormal behaviors of pedestrians in order to manage the crowd or prevent unexpected events. As Fig. \ref{fig:figure1}  shows, there are  many scenes (specific scenes)  equipped with  static cameras to record the events happened, followed by understanding the content of the captured videos and extracting useful information from the video sequences. In this process, pedestrian detection is the prerequisite for the subsequent event identification.

In recent years, with the improvement of computing power, machine learning,  especially deep learning \cite{NIPS2012_4824} has been playing a critical role in many topics of computer vision and makes a great progress, such as in image classification and object detection.

 \begin{figure*}[t]
	\centering
	\includegraphics[width=0.95\linewidth]{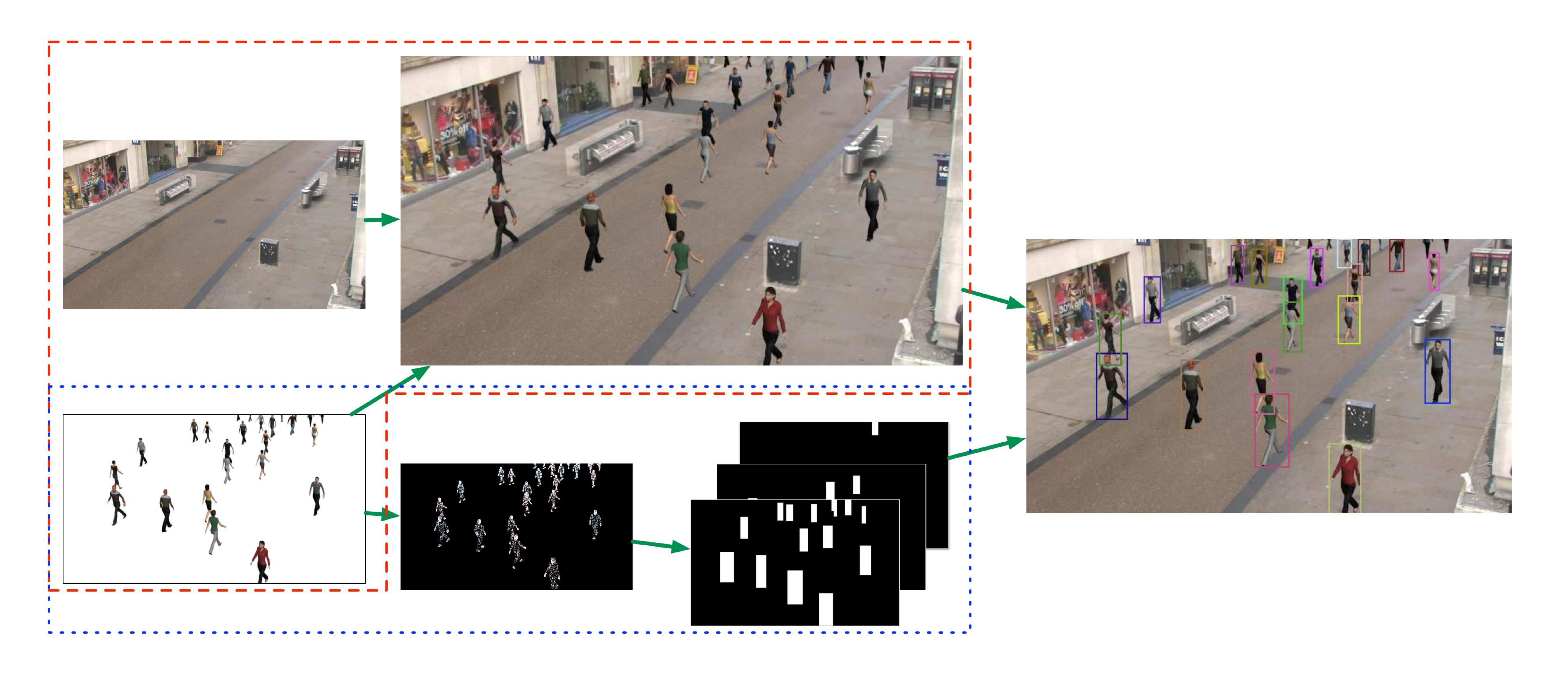}
    \captionsetup{font={footnotesize}}
	\caption{The process of rebuilding virtual scene and generating synthetic data  with accurate labeling information. \textbf{A. scene rebuilding and simulate pedestrians in scene(red box)}: (1) rebuild scene from geometry information (2) simulate pedestrians (3) generate   synthetic data from   virtual scene. \textbf{B. Labeling information Generated(blue box)}: (1) get the location information of pedestrian in photo through vertex render from 3d model. (2) obtain the final bounding box through computing.}
	\label{fig:screenshot001}
\end{figure*}

As a special object detection, pedestrian detection in generic scenes have achieved a great progress with large dataset trained 
 detectors , while the pedestrian detectors in generic scenes work bad when they are   directly used for pedestrian detection in specific scenes because of different geometry information, viewpoint, illuminations, resolutions, background and so on \cite{wang2011automatic} between specific scenes and generic scenes.

Although the generic pedestrian datasets presents   intra-class variations, each of the datasets still has its own inherent bias for the different methods collecting the data.
For example, since the INRIA \cite{dalal2005histograms} dataset is taken from the personal digital images collections, many of the people in the dataset are intent to make poses facing the cameras \cite{Htike2016Adapting}. This may be different from the natural pedestrian poses and actions from real-life situations.
For the dataset of Daimler and Caltech \cite{Dollar2012PAMI}, the pedestrians are captured from the on-board cameras from vehicles, with  different angles and view-points.

In fact,
the objective of labeled dataset is to provide the learned  classifier with large inner-class variations of pedestrian and non-pedestrian so that the resulting classifier can be generalizable to never-before-seen test dataset \cite{Htike2016Adapting}.
However, the pedestrians of generic scene and specific scenes with different angles, viewpoints and backgrounds are in different variable spaces, the pedestrian detectors in generic scenes works bad if they are used to the specific scenes directly.

For appearance-based pedestrian detectors, a large amount of data is important to train the appearance models.A  large amount of data happens to be lacked for the high cost to obtain.

It is necessary to adapt pedestrian detection models in generic  scenes to specific scenes in order to achieve a better result, so called `domain adaption' \cite{Htike2016Adapting}!

So far, there's much work done  in applying the specific scene pedestrian detector to generic scenes.
Wang \emph{et al.} \cite{wang2012transferring} uses transfer learning to transfer the generic pedestrian detectors to the specific scenes with little target labeled  data. Hottori at al. \cite{Hattori_2015_CVPR} make the 3d models as the proxy of the real world and train hundreds of classifiers in the scene of each possible position. Besides those, some researchers combine the synthetic and real data to train the specific pedestrian detector to overcome the lack of data.
Different from Hattori et al.'s idea, we directly rebuild the virtual world and simulate the pedestrian in the virtual scene with real background, and then collect a  large amount of labeled synthetic data.





In this paper, our main goal is to solve the problem of lacking labeled data in the specific scene based on the  parallel vision theory.

Parallel Vision theory was proposed by  Wang \emph{et al.} \cite{parallel_vision_wangkunfeng}, \cite{wang2016parallel}, \cite{wang2017parallelimaging} based on ACP (Artificial systems, Computational experiments, and Parallel execution) theory \cite{acp_theory}, \cite{wang2016steps}, \cite{liu2017paralleldata}, \cite{li2017parallel} attempting to solve the vision problems in the real world. Parallel Vision System follow the idea of "We can only understand what we create ". For parallel vision, A(CP) references to  building the virtual world corresponding to the related real world, then collecting  data from the virtual world; (A)C(P) references to obtain knowledge of the real world through Computing Experiments, finally applying the knowledge to both virtual world and real world to evaluate the effectiveness of the algorithm, namely Parallel Execution in (AC)P.



Based on the idea of parallel vision, we mainly  focus on building the virtual scene and generating a large amount of synthetic data from the virtual scene, and then we execute the Computational experiments with the generated data for pedestrian detection.With the virtual scene, we can generate much data for Computational experiments with labeling information, what's more, the virtual scene can free researchers from the annoying  process of preparing data  and make them  focus on algorithms.
The process of data generating is showed in Fig. \ref{fig:screenshot001}, more details can be found in Section 3.


 Our  contributions is proposing a method for training  pedestrian detectors in specific scenes when  new surveillance equipments are set up or   there is no   labeled data in a specific scene. And there are two purpose in this paper\ \textbf{(1)}: building virtual scene for specific scenes to generate synthetic data with labeling information and \ \textbf{(2)}: validating that the synthetic data is helpful to train a specific scene pedestrian detector, namely Computing Experiment in ACP theory.  

The remainder of this paper is organized as follows:Section II describes the related works. The virtual world are built, together with the process of generation of dataset and ground truth bounding boxes in  Section III. In Section IV, experimental results show that the synthetic data from virtual world is useful in training the pedestrian detectors in the real world dataset. At last but not the least, Section V is the conclusion.

\section{Related Work}
As  a part of ACP method,the virtual scene, also as virtual world  is important.
Changing the  motivation and appearance of target objects can generate a large mount of labeled data just as collecting from the real world! In such way, data-driven method can easily achieve the hoped goal.

As described by Bainbridge \cite{bainbridge2007scientific}, based on Video Game and Computer Game, the virtual world can simulate the complex physic world in vision and offers a new space for scientific research.
In fact, as part of parallel vision, the idea of using synthetic data generated from 3D models and virtual scenes for object detection is not new.

So far, there's much work been done based on the virtual world. Here, we mainly talk about some applications related to our work.


\textbf{Virtual Scene} In order to generate synthetic data, the first step is to build up the virtual scene. With the  development of GPU computing power, computer game  and 3d model, building up a virtual scene based game engine and 3d model software  like 3ds max, blender and so on becomes easy.
Prendinger \emph{et al.} \cite{Prendinger2013Tokyo} built Virtual Living Lab for traffic system simulation and analysis of driver behaviors based on OpenStreenMap, CityEngine and Unity 3D et al. Computer Game and 3d model softwares.
In the Virtual  Living Lab, they generated the traffic  road net  based on the free map data,and sense the context based on the interaction between the  agent  on board and the agent at the roadside. Karamouzas and Overmars \cite{Karamouzas2012Simulating} use the virtual scene to evaluate the proposed pedestrian motivation  model. The virtual scene is  also used for camera networks research \cite{Qureshi2007Smart,Starzyk2013Software}.

In our work, the main goal is to apply parallel vision concepts in specific scene, it is important to rebuild the same scene of  the specific scenes, so we directly  put variable pedestrians  in the image scene based on the technology of augment reality, which is similar to the the work done by Hattori \emph{et al.}  \cite{Hattori_2015_CVPR}, synthetic data generated based on the background of the specific scene.

\textbf{3D Model for Detection} For the similarities between real life object  and virtual object in  appearance and motivation, many researchers attempt to build a virtual proxy to stand for the real life object and achieve good results.
Brooks \cite{Brooks1981Symbolic}described a model-based vision system to model  image features.
Dhome \emph{et al.} \cite{Dhome1993Determination}presents a method to estimate the spatial attitude of articulated object from a single perspective image.
There's much work done in modeling human body shape \cite{Broggi2005Model}, pedestrian pose \cite{Heisele20163D}, hand gesture\cite{Molina2014A} and so on.
Hattori \emph{et al.} \cite{Hattori_2015_CVPR} use 3ds max model to train pedestrian detectors at each possible pixel in specific scene with static camera, and Marin \emph{et al.} \cite{marin2010learning} train a pedestrian detector in mobile scenario using virtual model.

In our artificial scene, we also model the pedestrians from 3ds Max. First, the model from modern 3d model software is similar to the real life pedestrians; what's more, with the synthetic data, we can easily  obtain large mount of ground-truth labeled data automatically.

\textbf{Scene Adaption}
In order to train a powerful model of pedestrian detection,generic data are collected through all possible way to cover all the possible  shape and appearance of pedestrian.However, all the existent dataset are not satisfy the principal and goal, for the variable scenes with variable viewpoint, lighting conditions and so on \cite{wang2012transferring}. In the specific scene, more effort needs to adapt the pedestrian model to a specific scene. So adapting the pre-trained models to a new domain has been an active area of research \cite{wang2012transferring,wang2014scene}.


Our work is also to adapting a generic pedestrian detector to a specific scene without real data, similar to their work \cite{Hattori_2015_CVPR}, but we are not train many detectors in each possible position. We train the generic pedestrian detector with synthetic data, and make sure that it is possible to use synthetic data generated from virtual world as a proxy of the real life.

\begin{figure}[t]
	\centering
	\includegraphics[width=0.95\linewidth]{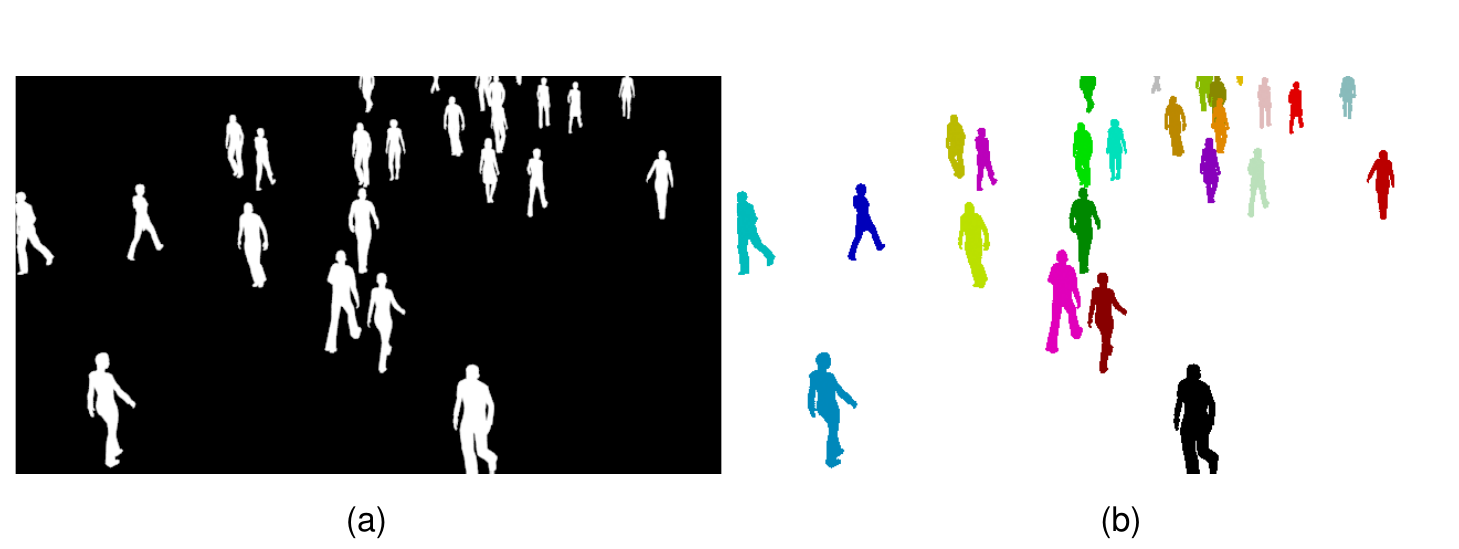}
    \captionsetup{font={footnotesize}}
	\caption{Output different formated image with solid color(black or white) background for following labeling work. \textbf{(a)} output with black-white object and background, \textbf{(b)} output with  different color for different pedestrian to show the overlapping relationship.}
	\label{fig:fig3}
\end{figure}

\section{Synthetic dataset And Methods}
Fig. \ref{fig:screenshot001} gives a pictorial illustration of the process of generating the synthetic dataset and labeling information from the virtual scene. And we evaluate the synthetic data with DPM \cite{Felzenszwalb2010Object-DPM} and Faster R-CNN \cite{Ren2015Faster-rcnn} methods.
 \subsection{Synthetic Dataset}
 In this paper, we use 3ds max to  rebuild the virtual scene with known  geometry information. With the virtual scene in 3D model, then we place the pedestrians in   target scene  and simulate pedestrians walking in the scene illustrated in red box of   Fig. \ref{fig:screenshot001}.


 \textbf{3D simulation}
 With the geometry information and background of the specific  scene,  we can easily rebuild the virtual scene, together with  the location of  obstacles in 3ds max. In the built virtual scene, we simulate 3d  pedestrians in the scene with  different appearance   walking around the virtual scene with arbitrary possible direction, walking speed and styles.


  In 3d simulation, we try our best to make the artificial scene photo realistic using the technology of AR. So we render  pedestrians walking  in different directions, walking around nearby, talking with each other within a group of two or three,  making  a call alone  and so on. With different configurations, we can easily generate variable data from virtual scene to learn a pedestrian model for the truth.

 \textbf{Ground-Truth Data} An important property of synthetic data is that we can conveniently obtain the ground truth data for the purpose of training the model and validating the model. In our work, we obtain the labeling information from   3d information of virtual scene directly! Although many researchers realize that it is possible to use synthetic data to train the real life model, and prove this method is effective in a way, they ignore the importance of generate the labels from the synthetic data.

 In  Sun \emph{et al.} \cite{sun2014virtual}, they generate synthetic  images that there is only one object in each  image. To get the labeling information, a parallel set of images was also generated which share the same rendering setting as the synthetic image generated for train the model except that the background is always white. The parallel generated images with white background are only used in automatically calculating the bounding box of non-white pixels, aka the target object bounding box in the image.
It is easy for only one object in the image, and many other works \cite{Peng2015Learning,Tzeng2015Towards} related with virtual object also generate labels  with the similar   method. The method is effective for one object in the image, but it does not work for images of multi-objects.

 For Multi-objects in the image, if the objects are not blocked each other, it is easy to label each object through search the connected t domain in the image \cite{di1999simple}. But there is no effective method for the blocked objects. It is impossible to obtain labels from image processing. At the first time, we also try the exist method to generate labels from the synthetic data, but we failed finally.



In the virtual world, we can easily obtain each parameter in the scene. Along with the render process, the points in the world $ X_W $ mapping to the camera coordinates $ X_C $ , then the image physics coordinates system $ X_{screen} $ and final in the image coordinates $X_{img}$ . The hole process can be display as equation.

 From world coordinate system to camera coordinate system:

 \begin{equation*}
 	 \begin{bmatrix}
 		x_c\\
 		y_c \\
 		z_c
 		
 	\end{bmatrix}=\begin{bmatrix}
 		R & t \\
 		0 & 1
 	\end{bmatrix}\begin{bmatrix}
 		x_w\\
 		y_w\\
 		z_w\\
 		1
 		
 	\end{bmatrix}=M_1\begin{bmatrix}
 		x_w\\
 		y_w\\
 		z_w\\
 		1
 		
 	\end{bmatrix}
 \end{equation*}

From camera coordinate system to physic coordinate system:
\begin{equation*}
	z_c\begin{bmatrix}
	u\\
	v
	\end{bmatrix}=\begin{bmatrix}
	f & 0 &0 \\
	0 & f &0 \\
	0 &0  &1
	\end{bmatrix}\begin{bmatrix}
	x_c\\
	y_c\\
	z_c
	\end{bmatrix}
\end{equation*}
Finally, convert the physics image system to image coordinate system:
\begin{equation*}
	\left\{\begin{matrix}
	u=x_u/d_x+u_0\\
	v=y_u/d_y+v_0
	\end{matrix}\right.
\end{equation*}

The final labeling information comes from the coordinates of image, the blocked relationship lies on the relationship of z-coordinates in camera.

In our work, we generate the synthetic data from virtual scene, meanwhile a parallel set of vertexes for each pedestrian is generated. From pedestrians' vertexes coordinates, we can easily obtain the labeling information showed as red box of Fig. \ref{fig:screenshot001}.


 \subsection{Object Detection Methods}
 The synthetic data is easy to obtain from 3d model together with labels. In order to validate the effectiveness of the synthetic data, we work on the data with Faster R-CNN \cite{Ren2015Faster-rcnn} and DPM (Deformable Part-based Model) \cite{Forsyth2014Object} pedestrian detection models, one for appearance-base method from CNN method  and one for edge-based method from HOG-SVM pedestrian detection.

 \textbf{DPM}\ The DPM (Deformable Part-based Model) model comes from PS (pictorial structure) \cite{Fischler1973The}, the relationship of different parts, and HOG (Histogram of Graphic) Feature \cite{dalal2005histograms}, the features of part characteristic, mainly describing the shape features of detected object and  achieve a great success    in object detection in the challenge of PASCAL VOC. The authors of DPM, Pedro Felzenszwalb \emph{et al.} were awarded a ``lifetime achievement" prize by the VOC organizers, which reflect the importance of DPM pedestrian detection method. The code can be found at \url{https://github.com/rbgirshick/voc-dpm}.

  \textbf{Faster R-CNN} Different from DPM model, Faster R-CNN \cite{Ren2015Faster-rcnn} is a appearance-based method coming from CNN.Recently, the CNN methods have achieve a great process in the vision problems as we mentioned above. As one instance of deep learning, Faster R-CNN, combining  Fast R-CNN \cite{DBLP:journals/corr/Girshick15} with RPN (Region Proposal Network), makes one stack object detection model and   performances well. The python code can be found at \url{https://github.com/rbgirshick/py-faster-rcnn}.

 DPM  and Faster R-CNN are two different  methods to describe the detected object and they   all perform well in object detection.
 We evaluate our proposed idea on both DPM and Faster R-CNN method, and the result show that the synthetic data is useful in object detection.

\section{Experimental Evaluation}

\subsection{Dataset}
\begin{figure}[t]
	\centering
	\includegraphics[width=0.95\linewidth,height=0.95\linewidth]{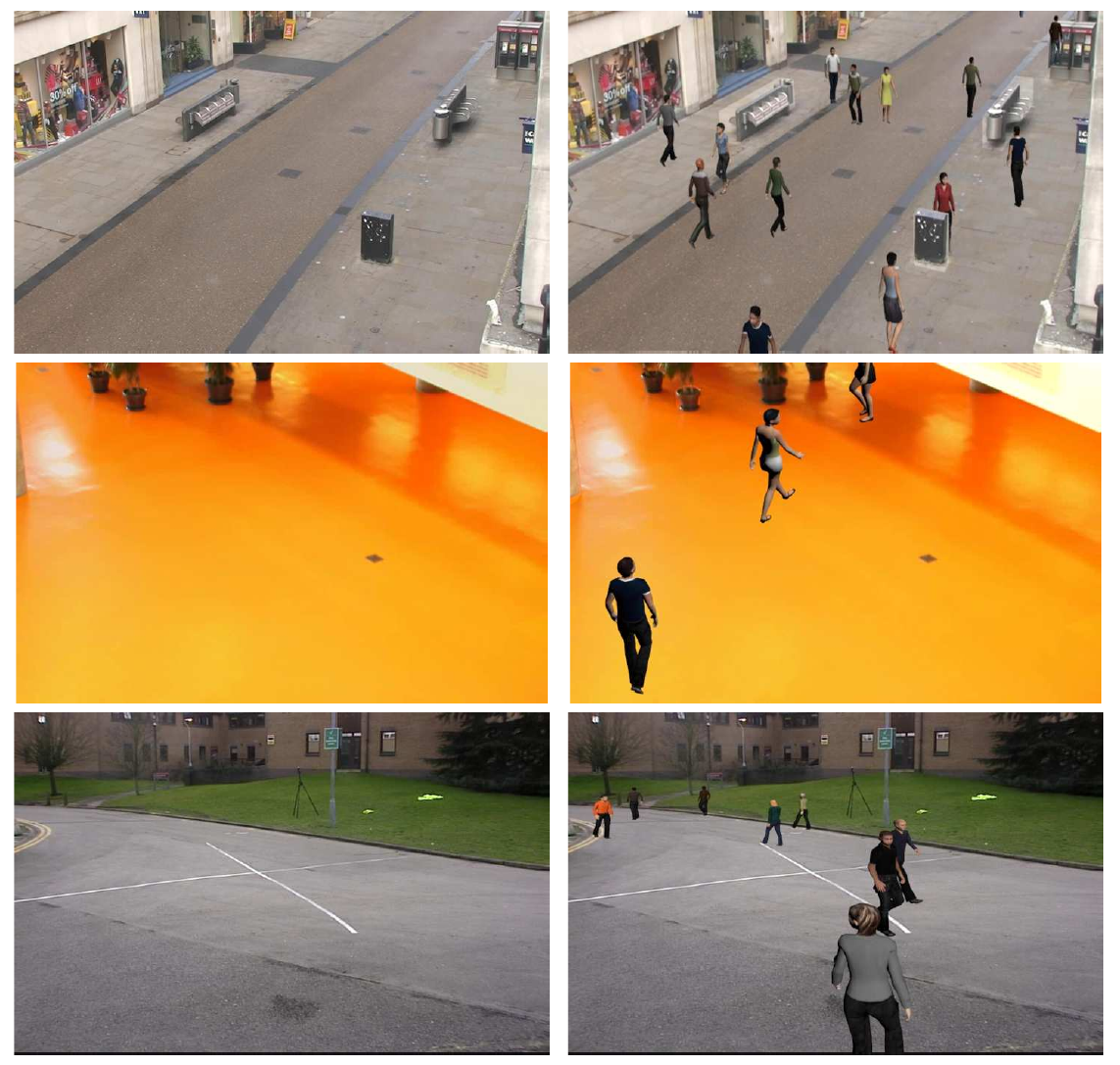}
\captionsetup{font={footnotesize}}
	\caption{
		The Evaluation Scenes with their corresponding geometric information and pedestrians simulated in them. Town Center (top), Atrium (middle) and PETS 2009 (bottom).
	}
	\label{fig:dataset}
\end{figure}

In order to validate the effectiveness of virtual scene for generating data in pedestrian detection, we evaluate it in different scenes as Fig. \ref{fig:dataset}.

\textbf{Town Center Dataset} \cite{benfold2011stable-Towncenter-dataset}: In our work, we mainly work  on the Town Center dataset, a video dataset of a semi-crowded town center with a resolution of 1920*1080 and a frame of 25 \textit{fps}. In order to consume the computing time and convenient to process, we also down-sample the videos to a standardized resolution of 640*360, together with the ground-truth labels as Hattori et al\cite{Hattori_2015_CVPR}. The original Town center Dataset only provides the validation set, and we use it for test dataset in our experiments.

\textbf{Atrium Dataset} \cite{Jodoin2014Urban}: Atrium pedestrian dataset was filmed at  École Polytechnique Montréal. It offers a view from the inside of the building and we can see pedestrians moving around crossing each others. This movie is intersecting since it allows to evaluate the performance on pedestrians. The movie resolution is 800x600. For the evaluation,   4540 frames (30 fps) of the movie were annotated.

\textbf{PETS 2009 Dataset} \cite{Ferryman2009An}: The PETS 2009 dataset consists of videos(at a resolution of 720X576) of  campus from    Kyushu University  including a number of pedestrians.While the dataset consists of videos captured  from 8 different views,  we just use a single 8th  camera view for our experiments. The ground truth information of pedestrians   are labeled by ourselves with  the tool labelimg from github \footnote{labelimg: \url{https://github.com/tzutalin/labelImg}}.

 \subsection{Baselines}

 We evaluate the efficiency of the virtual scenes and compare against with the following baselines.

\textbf{D-V/S}: D-V and D-S stand  for the DPM model of pedestrian detection trained with pascal voc 2007 dataset and synthetic dataset generated from the virtual world built by us respectively.

\textbf{F-V/S}: The same with D-V/S, F-V stands for the Faster R-CNN pedestrian detection model trained with dataset of person from pascal voc 2007 and the F-S stands for the Faster R-CNN model trained with the synthetic data generated from the dataset from the virtual scene.

All above are all have the same test set from the baseline of   specific scene pedestrian detection datasets: Town Center Atrium and PETS 2009. Except the experiments above, we also train and test Faster R-CNN and DPM model with person dataset  from pascal voc 2007.



 \subsection{Results}
 We compare   Faster R-CNN and DPM model to the baseline of  datasets of all selected specific scenes  based on pascal voc 2007  evaluation metric   for pedestrian detection.

\begin{figure*}[thb]
	\centering
	\subfloat[Town Center]{
		\includegraphics[width=6cm,height=4cm]{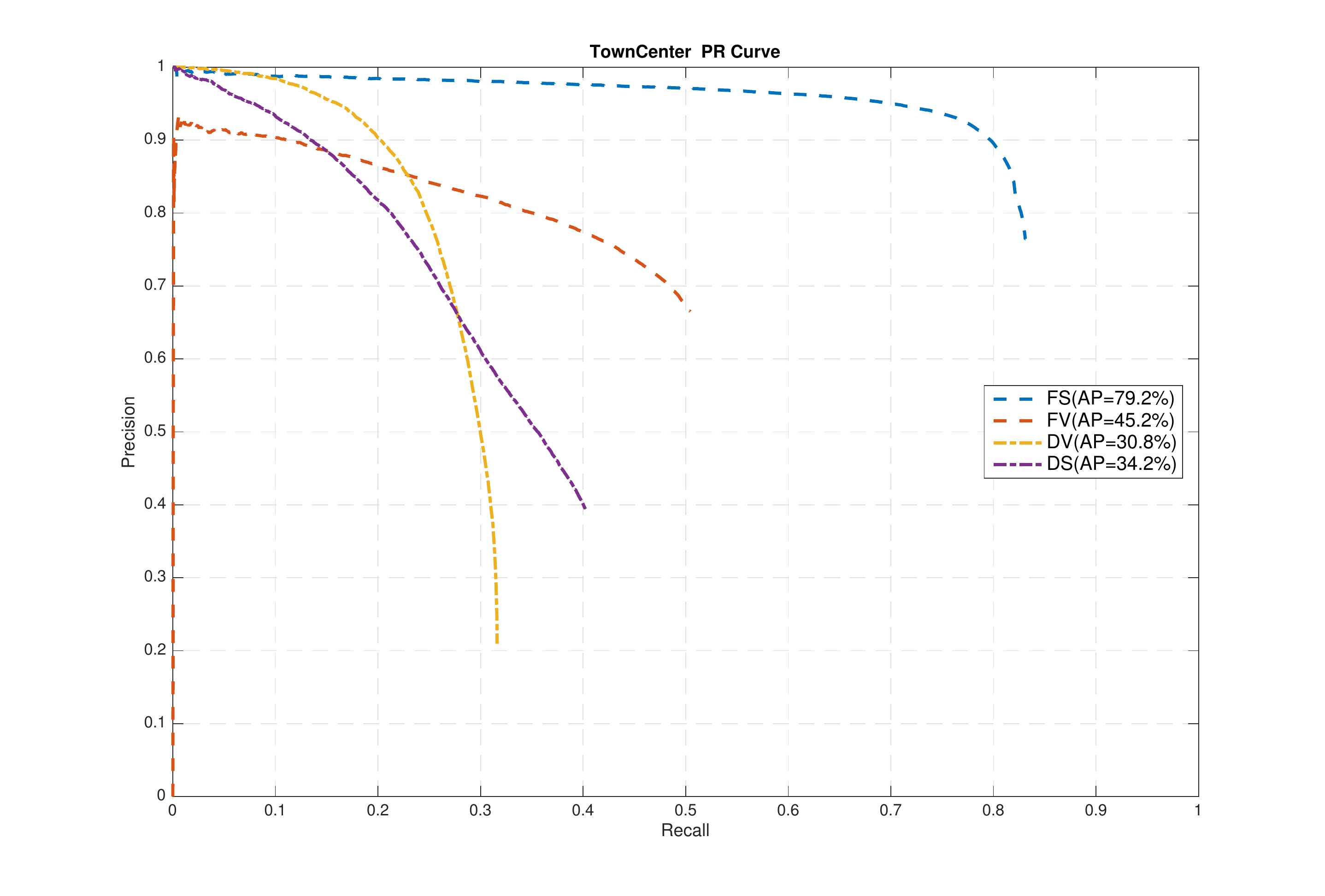}
	}
	\subfloat[Atrium]{
	\includegraphics[width=6cm,height=4cm ]{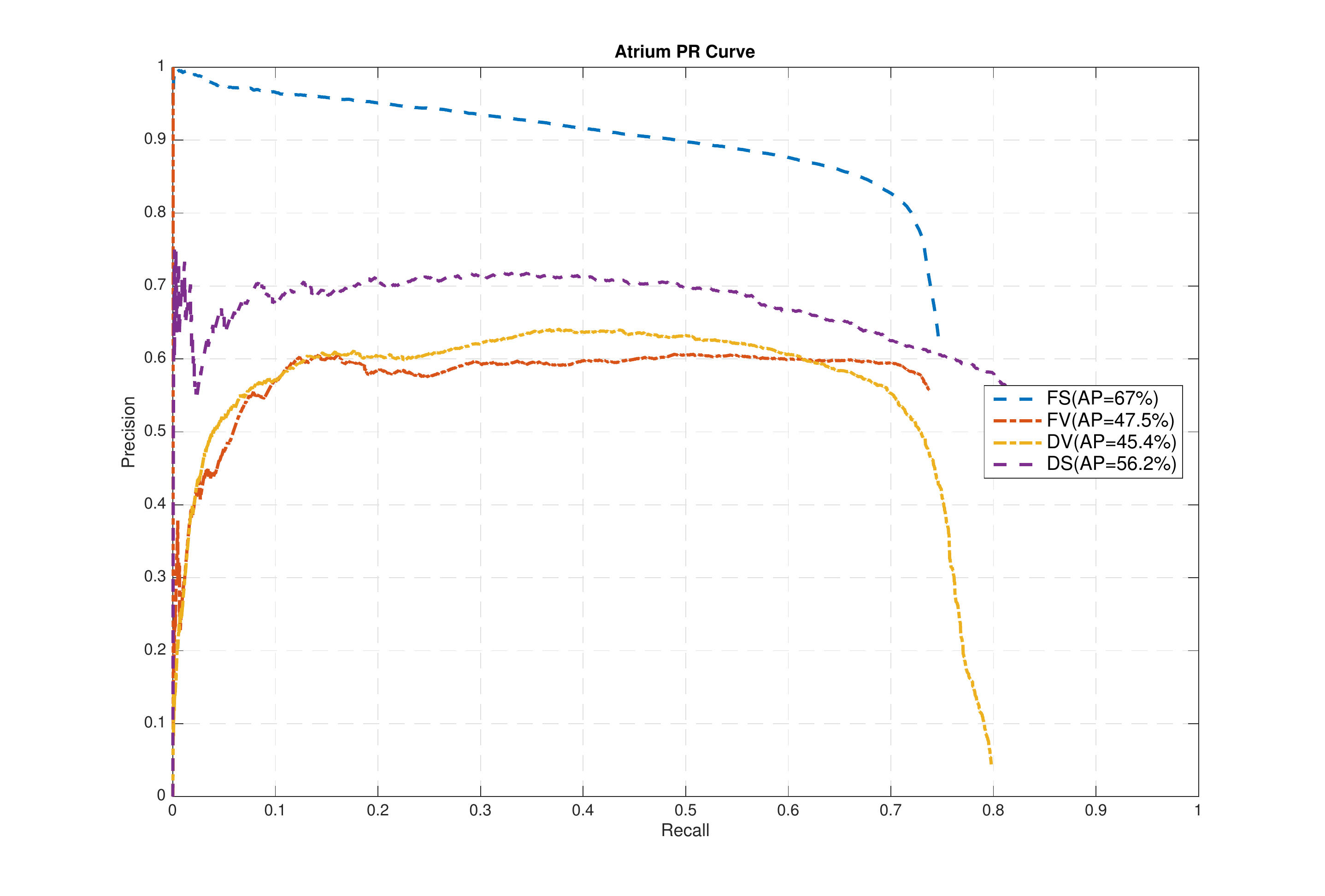}

}
	\subfloat[PETS 2009]{
	\includegraphics[ width=6cm,height=4cm]{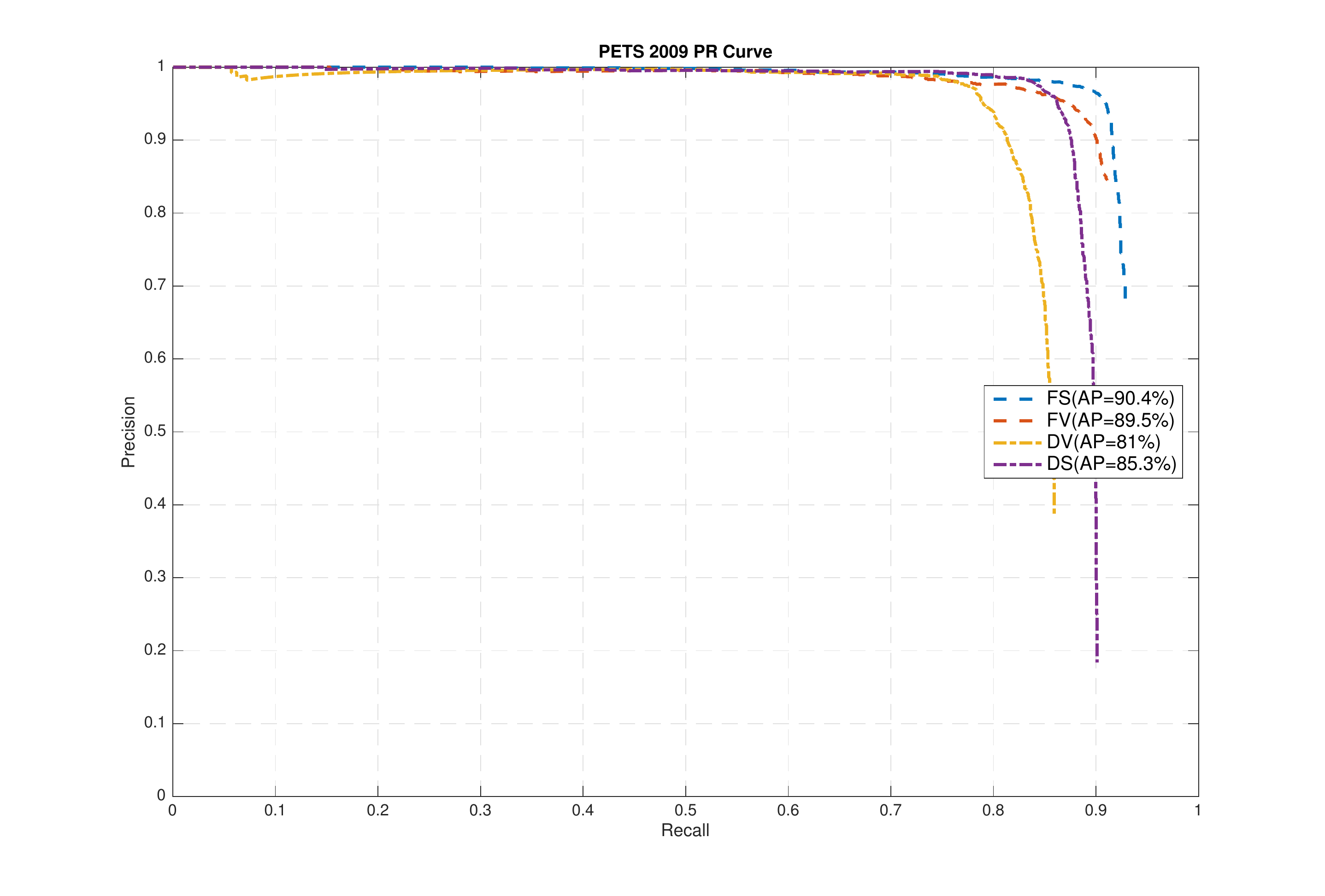}
}
\captionsetup{font={footnotesize}}
\caption{Precision-recall curve with PASCAL VOC 2007 metric in different scenes.}
\label{fig:pr-curve}
\end{figure*}

 The PR curves of pedestrian detection on the selected specific datasets are summarized in Fig. \ref{fig:pr-curve}, and the precision-recall curve show that the both Faster R-CNN and DPM model works well when trained with the   synthetic data generated from the virtual scene of Town center built by us than the models trained with general dataset of pascal voc 2007. From the AP of different method in different  scene evaluated datasets, we can conclude that in a specific scene, it is useful to train the model with synthetic data generated from virtual world.

 Table \ref{table-ap} shows the increment of AP with synthetic from virtual scene  compared to generic data of pascal voc 2007   in each evaluation scene for different method. In scene Town  Center and Atrium, the synthetic data improve the model much by a large margin, while the PEST 2009, for the pedestrians in the scene are much similar to the pedestrians in pascal voc, the generic model of DPM and Faster R-CNN have achieved  good results. However, the specific models  trained with synthetic data can still  achieve  results are not worse than them at least.

\begin{table*}[t]
	\centering
    \captionsetup{font={footnotesize}}
	\caption{AP Increment in each evaluation scene}
	\label{table-ap}
	\begin{tabularx}{18cm}{p{1.9cm}  X X X X X X X X X}
		\hline
	
		& \multicolumn{3}{c}{TownCenter} & \multicolumn{3}{c}{Atrium} & \multicolumn{3}{c}{PETS 2009} \\ \cline{2-10}
											& syn      & voc     & incre     & syn    & voc    & incre    & syn     & voc     & incre     \\ \hline
		Faster RCNN &      79.2\%    &    45.2\%     &  34\%         &  67\%      &  47.5\%       &  12.5\%        & 90.4\%        & 89.5\%        &  0.9\%         \\
		DPM          &     34.2\%     &   30.8\%      &     3.6\%      &    56.2\%    &  45.4\%      &   10.8\%       &   85.3\%      &  81\%       &   4.3\%        \\ \hline
	\end{tabularx}
\end{table*}

\section{Conclusion}
We propose a parallel vision approach to pedestrian detection by building the virtual world to generate a large amount of synthetic data and adapting the generic model to specific scenes. The experimental results show that synthetic data is able to learn a scene-specific detector without real labeled data. In this paper, the concerned objects are only pedestrians. But true scenes may contain pedestrians, cars, and trucks, making object detection more difficult. So in the future work, we will make the virtual world more complex and similar to the real scenes. Of course, the virtual world is useful not only for pedestrian detection, but also for other vision problems, such as segmentation, tracking, and pose estimation. We will do more computer vision research in virtual scenes and make the virtual world become a vision laboratory truly.





{
\footnotesize
 \bibliographystyle{IEEEtran}
 \bibliography{IEEEabrv,itsc17}
 }
\end{document}